\documentclass{article}

\usepackage[preprint]{corl_2026} 

\usepackage{graphicx}
\usepackage{xcolor}
\usepackage{svg}
\usepackage{booktabs}      
\usepackage{multirow}
\usepackage{array}
\usepackage[colorinlistoftodos,prependcaption,textsize=small]{todonotes}
\usepackage{amsmath}
\usepackage{hyperref}
\usepackage{amsfonts}
\usepackage{lipsum}
\usepackage{wrapfig}       

\definecolor{citecolor}{HTML}{0071BC}
\hypersetup{colorlinks,linkcolor={red},citecolor={citecolor}}

\title{AnyBody: Free-Form Whole-Body Humanoid Control from Arbitrary Keypoint Guidance}

\usepackage{authblk}
\author[1*]{Shuning Li}
\author[1]{Sikai Li}
\author[2]{Jiachen Li}
\author[1]{Mingyu Ding}

\affil[1]{University of North Carolina at Chapel Hill}
\affil[2]{Georgia Institute of Technology}
\begin{document}

\renewcommand{\thefootnote}{\fnsymbol{footnote}}
\footnotetext[1]{This work was done during an internship at UNC.}

\maketitle


\begin{figure}[h]
\centering
\vspace{-20pt}
\includegraphics[width=\linewidth]{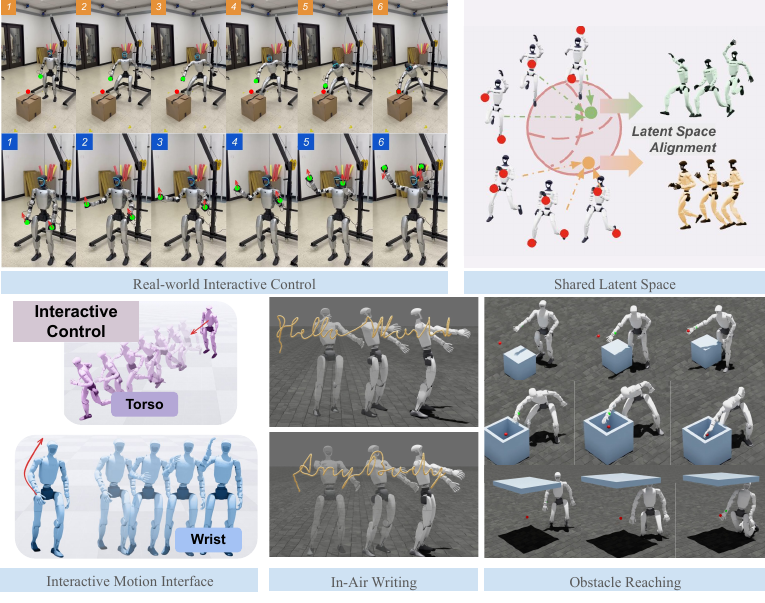}
\vspace{-16pt}
\caption{AnyBody learns a unified whole-body humanoid controller driven by arbitrary subsets of body keypoints. \textbf{Top left:} Real-world teleoperation examples on a Unitree G1 humanoid. Despite receiving only sparse keypoint commands, the policy synthesizes stable and coordinated whole-body motions. \textbf{Top right:} Diverse partial-body observations are mapped into a shared spherical latent motion space through latent-space alignment, enabling a single controller to operate under arbitrary keypoint configurations. \textbf{Bottom left:} Sparse keypoint trajectories provide an intuitive interface for interactive motion control, allowing users to specify behaviors through task-relevant keypoints such as the torso or wrist. \textbf{Bottom center and right:} 
We further expand the capability coverage of the controller via latent-space reinforcement learning toward downstream tasks, including in-air writing, obstacle reaching, and whole-body motion adaptation.}
\vspace{-8pt}
\label{fig:teaser}
\end{figure}

\begin{abstract}
We present AnyBody, a unified whole-body humanoid controller driven by an arbitrary subset of body keypoints chosen at deploy time. Prior physics-based trackers either rely on expensive full-body motion capture and error-prone trajectory retargeting, which bottleneck scalable data collection and policy learning, or decompose upper- and lower-body control into separate hierarchical representations, sacrificing the coordinated whole-body motions that loco-manipulation requires.
We close this gap by learning a single latent motion representation that any keypoint subset can address. To achieve this, we first train a privileged teacher tracker on a large unstructured motion corpus and distill it online into a deterministic encoder–decoder student whose latent space is a unit sphere. We then train a transformer keypoint encoder that admits any subset of body keypoints through masked self-attention, aligning it to the privileged latent. Additionally, we treat the frozen decoder as a motor prior and specialize downstream tasks with a lightweight residual corrector in the latent space. We demonstrate the effectiveness of AnyBody by tracking large-scale human motions from arbitrary keypoint subsets, free-form control, flexibly teleoperating, and learning downstream behaviors including locomotion, in-air writing, and obstacle-reach. Demos and code could be found at \url{https://hazel-hammer.github.io/anybody-project-page/}.
\end{abstract}


\section{Introduction}
\label{sec:intro}
Humanoid robots, sharing the general morphology of humans, hold strong potential for operating in human-centered environments and performing complex loco-manipulation tasks. However, controlling humanoids remains challenging due to their high degrees of freedom and the difficulty of maintaining stable whole-body coordination.

A dominant paradigm in recent humanoid control research~\cite{liao2025beyondmimic,seo2025fasttd3,huang2025towards,li2025bfm,kalaria2025dreamcontrol} is physics-based motion tracking. Policies first learn to imitate human motions in simulation, and are later controlled using dense motion references captured from mocap systems. While effective, this pipeline typically relies on expensive and cumbersome motion-capture systems, limiting scalable humanoid teleoperation and large-scale collection of whole-body control data.

To bridge this gap, recent works have explored hierarchical humanoid control frameworks that decouple upper-body manipulation and lower-body locomotion. By delegating locomotion to a pretrained lower-body controller~\cite{lu2025mobile,li2025hold,shi2026adversarial,li2025learning,amo,homie}, these methods allow users to teleoperate only the upper body through high-level commands. However, such methods typically assume a fixed control interface and explicitly separate upper- and lower-body control, limiting the coordinated whole-body behaviors required for loco-manipulation.

To enable more flexible forms of humanoid teleoperation while preserving coordinated whole-body behavior, we present AnyBody, a unified humanoid controller driven by an arbitrary subset of body keypoints. Rather than assuming a fixed control interface, AnyBody allows operators to specify motions using different keypoint configurations depending on the task and deployment setting, ranging from sparse upper-body tracking to full-body control or even single end-effector guidance. This flexibility enables intuitive and efficient humanoid control while allowing the robot to automatically synthesize physically plausible, robust, and human-like whole-body motions consistent with the provided keypoint constraints.

Our key idea is to learn a shared latent motion representation that unifies diverse forms of partial-body conditioning within a single whole-body control framework. We first learn a compact spherical latent space~\cite{xu2018spherical,fan2023unsupervised} from a large-scale human motion corpus~\cite{sonic}, such that diverse whole-body behaviors can be represented within a unified and reusable motion manifold. We then learn a transformer-based keypoint encoder that maps arbitrary subsets of body keypoints into the latent space through masked self-attention, enabling robust conditioning under varying keypoint configurations and missing observations.

We evaluate AnyBody across a diverse set of humanoid control scenarios. Quantitatively, our method accurately tracks sparse and varying keypoint subsets while maintaining smooth and coordinated whole-body motions comparable to joint-space tracking. We further demonstrate strong generalization to unseen human motions and show that our framework enables retargeting-free humanoid motion following directly from generated human keypoints, entirely bypassing conventional motion-retargeting pipelines. On real hardware, AnyBody supports flexible VR-based humanoid teleoperation under diverse control configurations, substantially simplifying the collection of loco-manipulation demonstrations. Finally, we demonstrate that the learned encoder-decoder representation captures reusable whole-body motion structure beyond pure tracking, enabling efficient downstream adaptation through latent-space reinforcement learning for tasks such as command-conditioned locomotion, obstacle reaching, and in-air writing.

Our contributions are threefold.
\textbf{1)} We introduce \textbf{AnyBody}, a unified humanoid control framework that supports
    arbitrary subsets of body keypoints at deployment time, enabling flexible and intuitive whole-body interactive control under diverse conditioning configurations.
\textbf{2)} We propose a shared spherical latent motion representation together with a transformer-based keypoint encoder that maps arbitrary sparse keypoint subsets into coordinated whole-body humanoid behaviors, enabling retargeting-free humanoid motion following, flexible real-world interactive control, and efficient downstream policy adaptation within a unified control framework.
\textbf{3)} We show that specifying task-relevant keypoint intent provides a
    flexible interface for synthesizing physically plausible humanoid motions, opening
    opportunities for scalable data synthesis and downstream policy
    learning.

\section{Related Works}
\label{sec:related works}
\noindent \textbf{Mocap-driven humanoid whole-body control.} Physics-based humanoid control has largely been built on imitating human motion capture. DeepMimic~\citep{deepmimic} established the phase-conditioned imitation reward that many subsequent trackers inherit, while Perpetual Humanoid Control~\citep{phc} scaled single-policy motion tracking to the full AMASS~\citep{mahmood2019amass} dataset in simulation. Recent works have further advanced large-scale motion tracking and sim-to-real transfer for humanoids through improved imitation objectives, larger motion corpora, more robust recovery mechanisms, and teacher--student distillation frameworks ~\citep{li2026coordex,dugar2025learning,han2025kungfubot2,lu2025mobile,shao2025langwbc,weng2025hdmi,xue2025leverb,wang2026experts,chen2025gmt}. Among them, SONIC~\citep{sonic} scales humanoid motion tracking along model size, dataset volume, and compute, achieving state-of-the-art tracking performance. However, these methods still rely on full-body motion capture as the control interface, which limits the scalability of humanoid loco-manipulation data collection.

\noindent \textbf{Humanoid control from sparse or arbitrary input.} A second line of work studies humanoid control from partial-body inputs. Human-to-humanoid teleoperation systems such as HumanPlus~\citep{humanplus}, OmniH2O~\citep{omnih2o}, and CLONE~\citep{clone} map sparse human signals, such as monocular body pose or VR head-and-hand tracking, to whole-body humanoid motion, with CLONE additionally introducing closed-loop drift correction. AMO~\citep{amo} expands the reachable workspace through hybrid trajectory optimization, while HOMIE~\citep{homie} and related decoupled frameworks separate upper- and lower-body control into independent policies driven by dedicated interfaces. Closest to our setting, the Masked Humanoid Controller~\citep{mhc} and HOVER~\citep{hover} admit partially specified targets through masked state variables or predefined command modes. However, existing approaches typically either assume a fixed control interface, rely on predefined command structures, or decouple whole-body coordination into separate control modules.
\section{Method}
\label{sec:method}

\textbf{Notation.} The humanoid has proprioceptive state $s^p$ and a goal command $s^g$
derived from the reference motion; the policy outputs an action $a\in\mathbb{R}^{n_a}$ (PD
targets for the $n_a$ actuated joints). $E$ denotes an encoder into a latent
$z\in\mathbb{R}^{d_z}$, $D$ the decoder, and $\pi^{T}$ a privileged teacher. Keypoints are a
set of $N$ body points; a binary mask $m\in\{0,1\}^{N}$ selects the commanded subset
and $V=\{i:m_i=1\}$ is the visible set. $\mathrm{sg}[\cdot]$ is the stop-gradient operator and
$\widehat{(\cdot)}$ denotes $\ell_2$-normalization.

\begin{figure}[t]
\centering
\includegraphics[width=\linewidth]{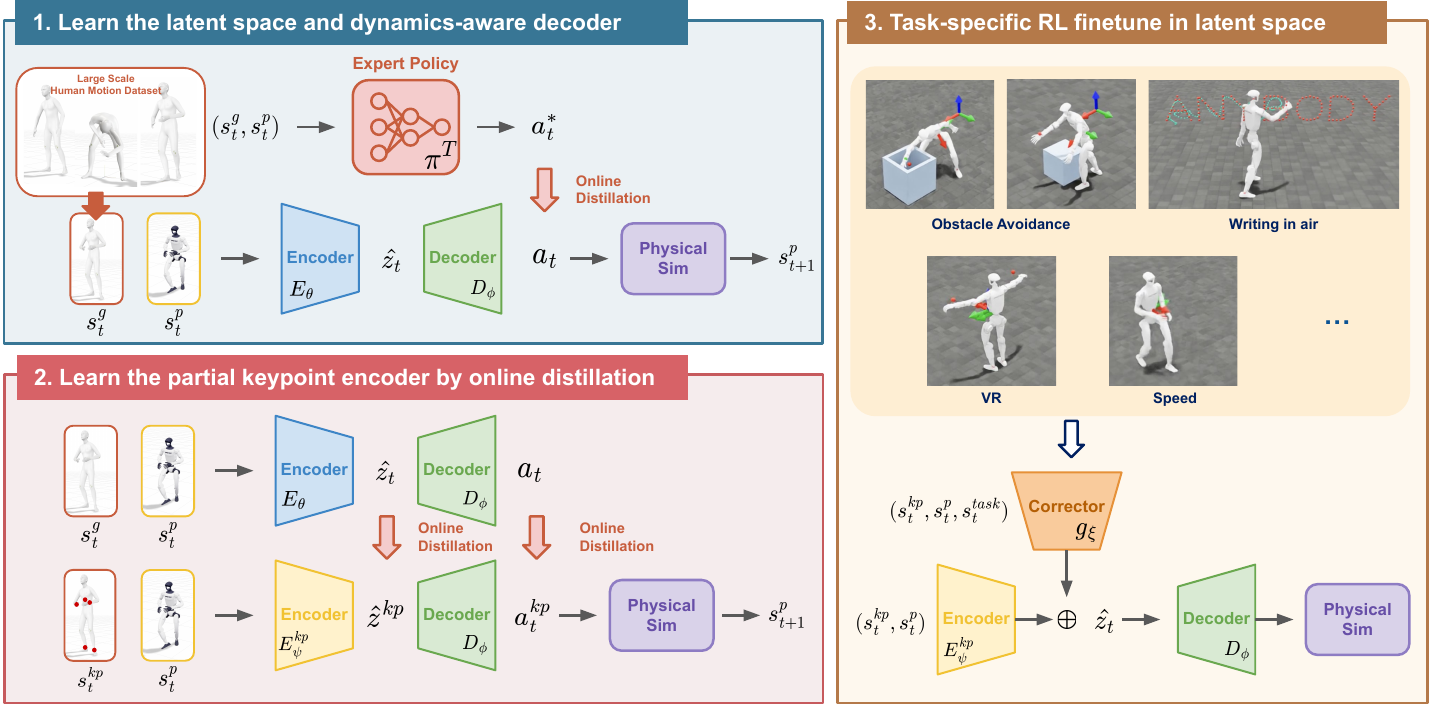}
\vspace{-16pt}
\caption{Overview of AnyBody. (a) We first train a privileged teacher tracker on a large-scale motion corpus and distill it into a deterministic encoder--decoder, yielding a compact spherical latent motion representation and a dynamics-aware decoder. (b) With the latent space and decoder frozen, we train a transformer-based keypoint encoder by online latent distillation, aligning arbitrary sparse keypoint observations with the privileged latent and enabling control from varying keypoint subsets. (c) We reuse the frozen decoder as a motor prior and specialize the controller through residual reinforcement learning in the latent space, expanding capability coverage toward downstream tasks such as obstacle reaching, in-air writing, and real-world interactive control.
\vspace{-12pt}
}
\label{fig:method}
\end{figure}

\textbf{Overview.} AnyBody is trained in three stages (Figure.~\ref{fig:method}). Stage~1
distills a privileged teacher into a deterministic encoder--decoder, yielding a compact latent
space and a dynamics-aware decoder (\S\ref{sec:method:latent}). Stage~2 freezes the decoder
and trains a transformer keypoint encoder by online latent distillation, so the policy can be 
driven by arbitrary keypoint subsets (\S\ref{sec:method:kp}). Stage~3 specializes the tracker
in downstream tasks via residual RL in the latent space, with the frozen decoder as a motor
prior (\S\ref{sec:method:rl}).

\subsection{Learning the latent space and dynamics-aware decoder}
\label{sec:method:latent}
\textbf{Latent motion representation.} Rather than learning motor skills directly in the action 
space, we first compress them into a compact latent space and map latents to actions with a 
prior-enriched decoder. Following common practice~\cite{hu2023teacher,yamada2024twist,wang2021knowledge,zhang2025distillation,yang2025multi}, a privileged teacher $\pi^{T}$ is trained to track a large-scale 
motion dataset and distilled online into a deterministic encoder $E_\theta$ and decoder $D_\phi$. We project the latent onto the unit
sphere, so the representation lives on $\mathbb{S}^{d_z-1}$:
\begin{equation}
  z = \widehat{E}_\theta(s^g, s^p) \in \mathbb{S}^{d_z-1}, \qquad
  a = D_\phi(z, s^p)
\end{equation}
The student minimizes a behavior-cloning~\cite{torabi2018behavioral} term against
the teacher and a cosine smoothness regularizer on consecutive latents~\cite{luo2024universal},
\begin{equation}
  \mathcal{L}_{1} = \big\| D_\phi(z, s^p) - \pi^{T}(s) \big\|_2^2
  \;+\; \lambda_{\mathrm{sm}}\big(1 - \cos(z_t, z_{t-1})\big).
\end{equation}
With access to simulation and online distillation, 
the deterministic student matches the teacher and can serve as the latent-space teacher for Stage~2, 
without any observed loss of latent smoothness or downstream usability.

\subsection{Learning the partial keypoint encoder by online distillation}
\label{sec:method:kp}
\textbf{Keypoint encoder.} We replace the joint-command encoder with a keypoint encoder
$E^{kp}_\psi$ that consumes a set of masked keypoints and the proprioceptive state and emits a
latent in the same space. $E^{kp}_\psi$ is a self-attention transformer over per-keypoint
tokens; absent keypoints are dropped via the attention mask $m$, so any subset is admissible
at test time. Since anchor-frame keypoint positions carry little velocity information, each 
token stacks a $0.5$\ s window of history and future ($15$ frames), giving the encoder the 
temporal context to infer motion.

\textbf{Latent distillation.} With the Stage-1 encoder $E$ and shared decoder $D$ frozen, we
align the keypoint latent to the privileged joint-command latent on the unit sphere, and add an 
action-space alignment that further anchors the decoded action,
\begin{equation}
\mathcal{L}_2 = \big(1 - \cos(\hat z^{kp},\, \mathrm{sg}[\hat z])\big)
  \;+\; \lambda_a \big\| D(\hat z^{kp}, s^p) - \mathrm{sg}[D(\hat z, s^p)] \big\|_2^2,
\end{equation}
where $\hat z^{kp} = E^{kp}_\psi(k\odot m, s^p)$ and $\hat z = E(s^g, s^p)$. Training follows a three-phase masking
curriculum: all keypoints visible; per-keypoint Bernoulli masking with keep-probability
annealed $1.0\!\to\!0.4$; then sampling semantic deployment modes (torso-, wrists-, ankles-only, etc.).

\subsection{Task-specific RL finetune in latent space}
\label{sec:method:rl}
\textbf{Residual latent-space RL.} To specialize the tracker, we finetune \emph{in the latent
space} with PPO~\cite{schulman2017proximal}: the RL action is the latent $z$ and the environment applies $a = D(z, s^p)$ with $D$
frozen as a motor prior. Rather than finetuning $E^{kp}_\psi$ or inserting LoRA adapters~\cite{zhang2023lora}, we
learn a small residual corrector~\cite{silver2018residual,johannink2019residual,alakuijala2021residual,zhang2019deep} $g_\xi$,
\begin{equation}
  z = \widehat{\,\mu + \alpha\,\Delta z\,}, \qquad
  \Delta z = g_\xi(o, \mu), \qquad
  \mu = E^{kp}_\psi(k\odot m, s^p),
\end{equation}
with small-gain initialization ($\Delta z \!\approx\! 0$) so the policy reproduces the
distilled tracker at initialization---a safe start that preserves the prior while training few parameters.
\section{Experiments}
\label{sec:experiments}

We evaluate AnyBody across a diverse set of simulated and real-world humanoid control scenarios. Specifically, our experiments are designed to answer four key questions:
\textbf{(Q1)} Can AnyBody accurately track sparse and varying keypoint subsets while preserving smooth and coordinated whole-body behavior?
\textbf{(Q2)} How well does the learned controller generalize to unseen human motions and diverse motion categories?
\textbf{(Q3)} Does AnyBody support practical real-world humanoid teleoperation under flexible control configurations?
\textbf{(Q4)} Does the learned latent space capture the rich diversity of whole-body motions, and can latent-space reinforcement learning further expand the capability coverage of AnyBody for downstream tasks?

To answer these questions, we first perform quantitative tracking evaluations under different keypoint-commanding modes (\S\ref{sec:Q1}), followed by a large-scale generalization analysis on unseen human motions (\S\ref{sec:Q2}). We then validate the framework on real-world VR teleoperation tasks (\S\ref{sec:Q3}). Finally, we study latent-space reinforcement learning for downstream humanoid behaviors including command-conditioned locomotion, obstacle reaching, and in-air writing (\S\ref{sec:Q4}).

\subsection{Quantitative Analysis (Q1)}
\label{sec:Q1}

\begin{table}[t]
\centering
\caption{\textbf{Quantitative tracking performance under different keypoint-commanding modes.}
We evaluate tracking success rate (SR), point-of-interest (POI) position error, and POI velocity error across diverse sparse-conditioning configurations. Our method maintains high tracking success and smooth whole-body motion even under highly sparse keypoint commands.}
\label{tab:q1_tracking}
\resizebox{\linewidth}{!}{
\begin{tabular}{lcccc}
\toprule
\textbf{Command Mode} & \textbf{\# Points} & \textbf{SR (\%)} $\uparrow$ & \textbf{POI Pos. Err. (cm)} $\downarrow$ & \textbf{POI Vel. Err.} $\downarrow$ \\
\midrule
Full Body                         & 5  & 97.6 & 10.90 & 0.301 \\
Upper Body (Wrists + Torso)      & 3  & 94.8 & 9.48  & 0.284 \\
Wrists Only                      & 2  & 94.3 & 10.41 & 0.325 \\
Single Wrist (L/R)               & 1  & 93.7 & 10.79 & 0.335 \\
Torso Only                       & 1  & 94.3 & 8.34  & 0.216 \\
Ankles Only                      & 2  & 95.4 & 13.39 & 0.351 \\
\midrule
Joint Command w/ Privileged Info & -- & 99.3 & 8.31  & 0.225 \\
Joint Command w/o Privileged Info& -- & 99.1 & 16.98 & 0.378 \\
\bottomrule
\end{tabular}
}
\end{table}

We train AnyBody on a large-scale open-source human motion dataset
containing diverse whole-body behaviors.
Since our primary focus is humanoid loco-manipulation, we deliberately
remove motion categories that are less relevant to practical
manipulation settings, such as crawling and highly dynamic jumping
motions, and train on the remaining motion corpus.

Table~\ref{tab:q1_tracking} reports quantitative tracking performance
under different keypoint masking configurations.
AnyBody maintains consistently high tracking success across all
sparse-conditioning modes, including upper-body-only, single-wrist,
torso-only, and ankle-only control.
Even under highly under-constrained settings such as single-wrist
tracking, the policy preserves stable behavior with success rates
above $93\%$.
Notably, the average POI position error remains consistently around
$10$~cm across most masking modes despite large variation in the
number and location of commanded keypoints, indicating that AnyBody
can robustly follow sparse user guidance while maintaining coordinated
whole-body motion.

For reference, we additionally report the performance of joint-space
tracking policies trained with and without privileged anchor-position
observations.
We note that these comparisons are not strictly equivalent: AnyBody
receives keypoint positions expressed in the robot anchor frame, while
conventional joint-space trackers typically do not observe explicit
global position information.
We therefore include the last two rows primarily as references
rather than exact apples-to-apples comparisons.

\subsection{Open-Ended Generative Analysis (Q2)}
\label{sec:Q2}
\begin{figure}[t]
  \centering
  \includegraphics[width=\linewidth]{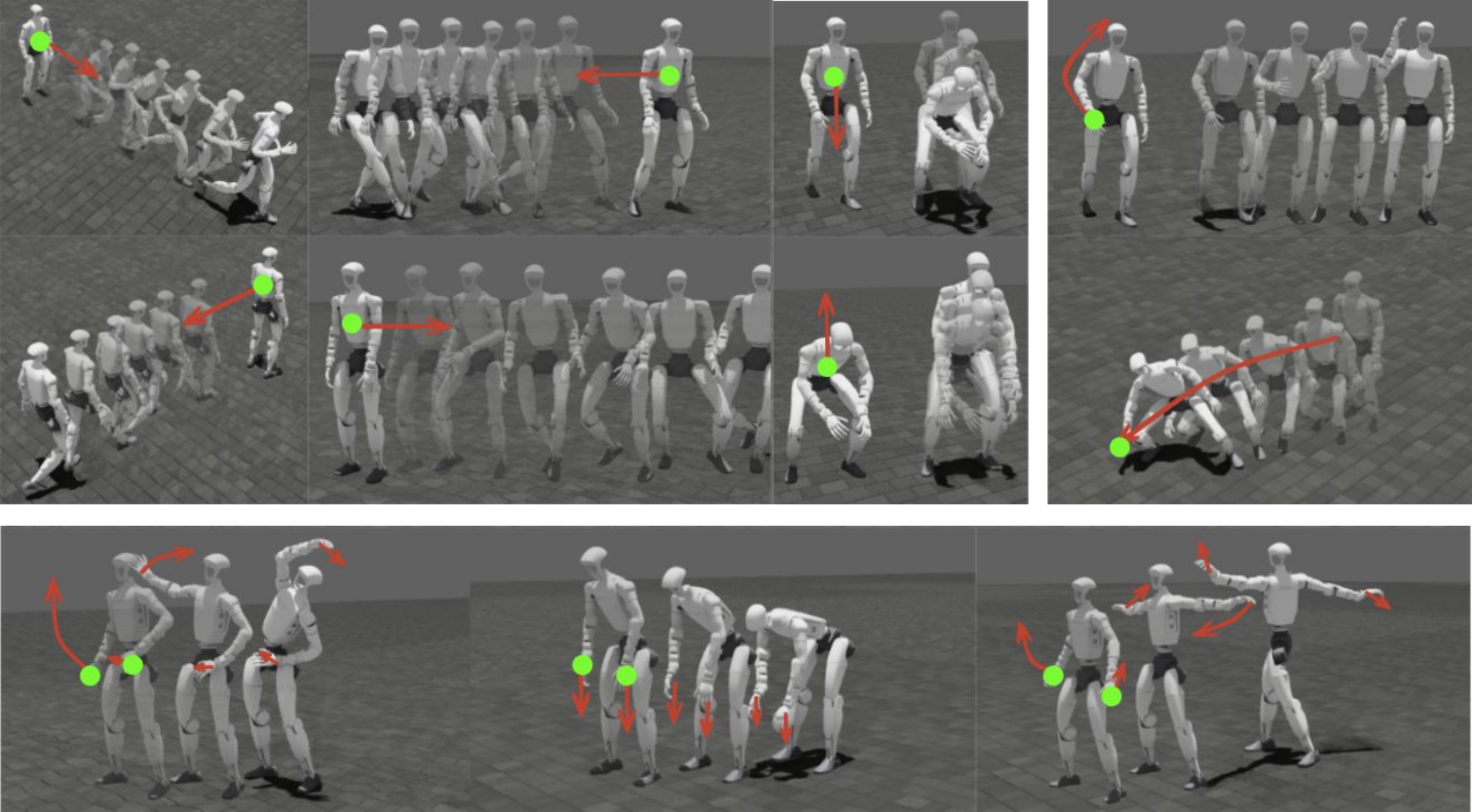}
  \caption{\textbf{Open-ended behavior generation from sparse keypoint commands.}
     Red arrows denote manually specified keypoint trajectories. AnyBody follows diverse sparse commands while synthesizing physically plausible and coordinated whole-body motions, including single-keypoint control: directional locomotion, arm swings, arm raises, bending, squatting, and multi-keypoint/closed-loop control. These examples illustrate that a small set of task-relevant keypoint trajectories is sufficient to induce a wide variety of whole-body behaviors through the learned latent motion representation.
}
  \label{fig:fig_simple}
  \vspace{-12pt}
\end{figure}

\begin{wrapfigure}{r}{0.25\linewidth}
\vspace{-\baselineskip}
\centering
\includegraphics[width=\linewidth]{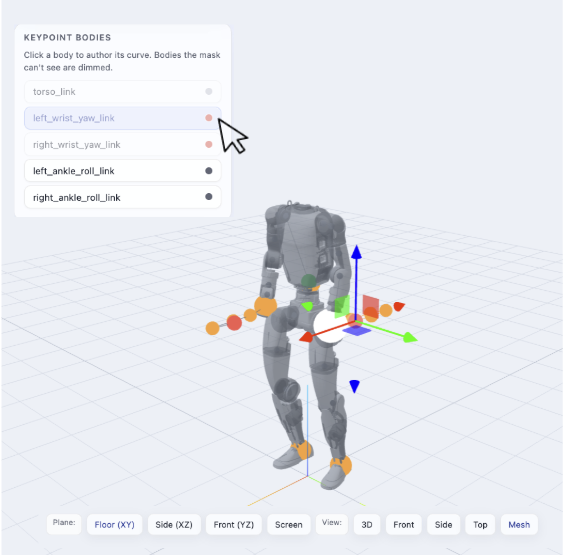}
\vspace{-16pt}
\caption{Lightweight traj-creation interface.}
\vspace{-20pt}
\label{fig:player}
\end{wrapfigure}

Beyond quantitative benchmarks, we further study the open-ended controllability and generative capability of AnyBody under manually specified keypoint trajectories. Intuitively, the partial keypoint tracker can robustly follow a wide range of simple sparse commands while synthesizing coordinated whole-body behaviors, as shown in Figure \ref{fig:fig_simple}. 

To facilitate interactive evaluation, we additionally develop a lightweight trajectory-creation interface, shown in Figure \ref{fig:player}, which allows users to manually design sparse keypoint trajectories and export them as trackable motion files. This interface enables rapid qualitative probing of the capability boundary of the learned latent motion representation.

Despite the broad coverage of the pretrained tracker, we observe several failure modes. In particular, the policy struggles when commanded trajectories correspond to poses rarely observed in the training corpus, such as extremely low reaching motions, or when the target trajectories are substantially out-of-distribution, such as large-scale in-air handwriting trajectories. As discussed in \ref{sec:downstream RL}, we show that many of these failure cases can be substantially improved through lightweight downstream latent-space RL finetuning. 

\subsection{Real-robot experiment (Q3)}
\label{sec:Q3}
\begin{wrapfigure}{r}{0.50\linewidth}
\vspace{-20pt}
\centering
\includegraphics[width=\linewidth]{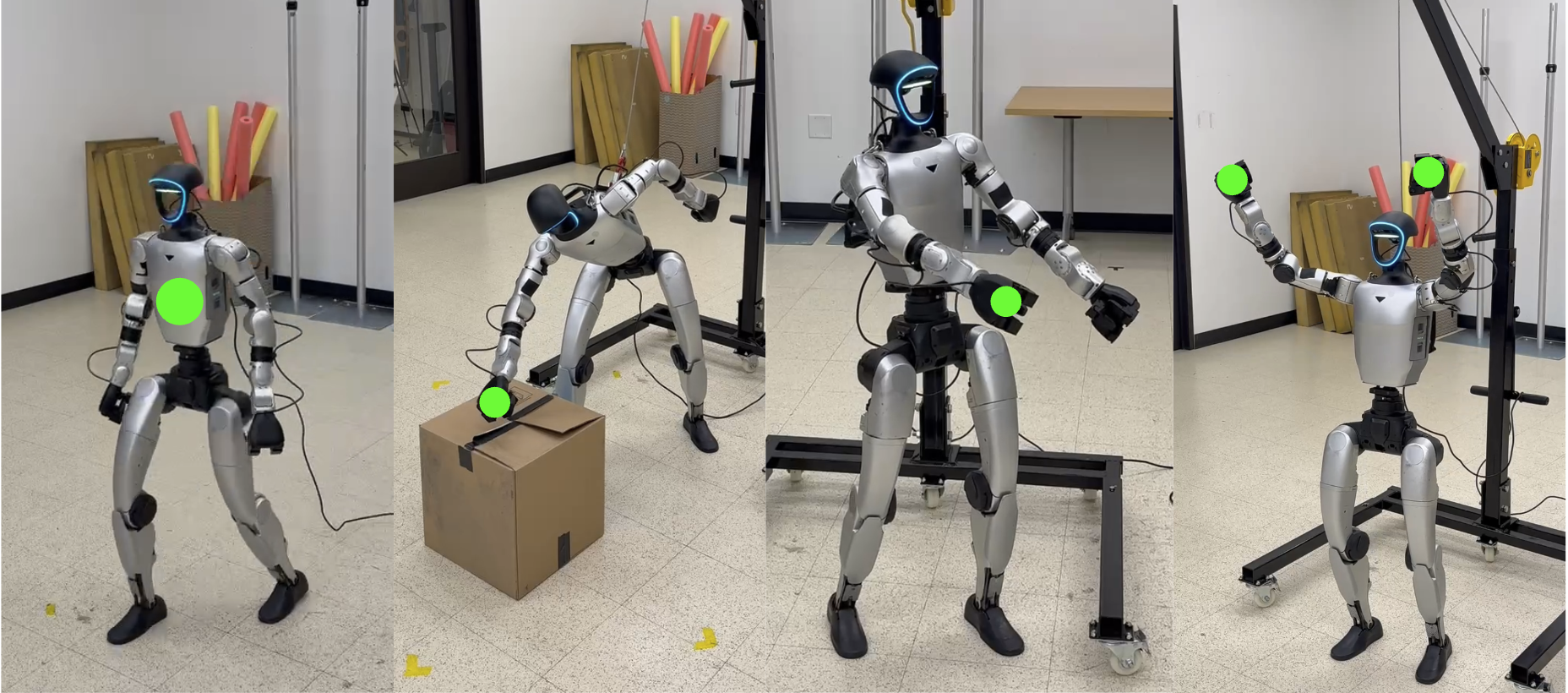}
\vspace{-20pt}
\caption{Real-world experiment overview. More demos could be found in the Appendix.}
\vspace{-12pt}
\label{fig:realworld experiment}
\end{wrapfigure}

We further evaluate whether AnyBody transfers from simulation to real-world humanoid teleoperation. Our hardware setup consists of a Unitree G1 humanoid together with an Apple Vision Pro headset for sparse keypoint capture and teleoperation input.

We study whether sparse keypoint commands can effectively drive the humanoid policy while preserving stable and coordinated whole-body behavior. In particular, we evaluate VR teleoperation using one or both wrist keypoints under different control configurations. Despite receiving only sparse upper-body guidance, the policy is able to synthesize coordinated full-body motions that remain stable during real-world execution. Representative teleoperation results are shown in Figure~\ref{fig:realworld experiment}. More detailed visualizations can be found in the Appendix.

\subsection{Downstream RL finetune (Q4)}
\label{sec:downstream RL}
\label{sec:Q4}
\begin{figure}[t]
  \centering
  \includegraphics[width=\linewidth]{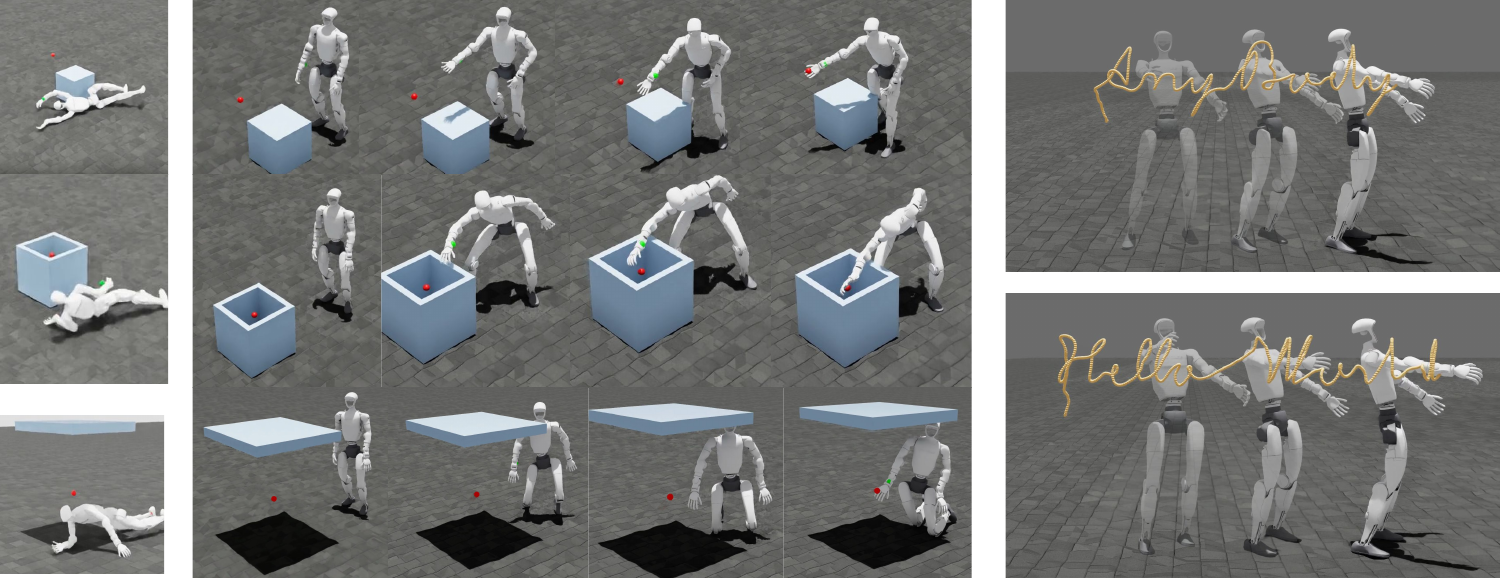}
  \caption{\textbf{Latent-space RL expands AnyBody's capability coverage.}
    Latent-space RL expands AnyBody's capability coverage beyond direct motion following. Left: when driven solely by sparse keypoint tracking objectives, the pretrained controller naively follows the commanded trajectory and may fail in task-constrained environments, for example by colliding with obstacles. Middle: latent-space RL adapts the motion prior to discover task-aware whole-body strategies, including leaning, bending, squatting, and reaching into containers while maintaining stable humanoid motion. Right: the same framework enables fine-grained wrist trajectory tracking and in-air writing. These results demonstrate that the learned latent representation serves as an effective motion prior that can be efficiently specialized to downstream tasks through reinforcement learning.
 }
  \label{fig:downstream_rl}
\end{figure}

While AnyBody may fail to accurately follow keypoint trajectories that lie
outside the distribution of the training motion corpus, the learned latent
space already captures a rich distribution of physically plausible whole-body
motions. We show that reinforcement learning in the latent space can further
expand the capability coverage of AnyBody by adapting the pretrained motion
prior toward task-specific objectives. We design three downstream tasks to
evaluate whether latent-space RL can expand the capability coverage of AnyBody
beyond the original motion dataset while preserving coordinated whole-body
behavior.

\textbf{Obstacle-reach.}
The humanoid must move its right wrist to a fixed world-frame target while
avoiding obstacles that block the direct reaching path. Successfully reaching
the target requires the policy to adopt different whole-body motion strategies
such as bending, leaning, or squatting, testing whether the latent policy can
exploit the multi-modal structure of the motion prior under task constraints.

\textbf{Wrist-writing.}
The humanoid traces in-air polylines with its right wrist to spell randomly
sampled words at varying scales and torso-relative locations. For large
writing trajectories, the policy must additionally coordinate locomotion and
upper-body leaning to bring the wrist within reachable workspace. This task
evaluates whether latent-space RL can synthesize fine-grained end-effector
trajectories while maintaining globally coordinated whole-body motion.

\begin{table}[t]
  \centering
  \small
  \caption{Downstream-task success rates with and without latent-space PPO
    finetuning. The frozen motion prior already solves a meaningful fraction
    of episodes; latent-space RL pushes all tasks above 95\%.}
  \begin{tabular}{lcc}
    \toprule
    Task & SR w/o RL finetune & SR w/ RL finetune \\
    \midrule
    \multicolumn{3}{l}{\emph{Obstacle-reach}} \\
    \quad Open               & 54.68\% & 97.09\% \\
    \quad Barrier            & 10.49\% & 96.04\% \\
    \quad Low-clearance      & 44.36\% & 99.47\% \\
    \quad Container          & \phantom{0}3.04\% & 95.56\% \\
    \midrule
    Wrist-writing            & \phantom{0}0.00\% & 97.87\% \\
    \bottomrule
  \end{tabular}
  \label{tab:rl_results}
\end{table}

Task success is defined as maintaining the commanded point-of-interest (POI)
within a 5\,cm tracking error throughout the trajectory for wrist-writing and 
reaching the target POI within 2\,cm at
the end of the episode while maintaining a stable and coordinated whole-body
behavior for obstacle-reach.

As shown in Table~\ref{tab:rl_results}, the policy adapts effectively to all
downstream tasks, suggesting that the learned latent space already captures a
rich manifold of coordinated humanoid motions and that the pretrained masked
keypoint encoder provides a strong initialization for downstream optimization.
Furthermore, latent-space RL converges rapidly and stably across tasks,
indicating that reinforcement learning primarily composes and adapts existing
motion modes rather than learning whole-body behaviors from scratch. 
\section{Conclusion}
\label{sec:conclusion}
We presented AnyBody, a unified humanoid control framework that enables whole-body motion generation from arbitrary subsets of body keypoints. By learning a shared latent motion representation together with a masked transformer keypoint encoder, AnyBody supports flexible sparse-conditioning interfaces while preserving coordinated and physically plausible whole-body behavior. We hope this work provides a step toward more flexible, scalable, and general-purpose humanoid control systems that bridge motion teleoperation, imitation and autonomous skill learning.

\noindent \textbf{Limitation and Future Works.}
\label{sec:limitations}
Although AnyBody demonstrates strong performance across both simulation and real-world experiments, maintaining high tracking accuracy across a broad range of motion distributions and adapting efficiently to downstream tasks, several limitations remain. First, the controller still struggles on certain edge-case motions due to limited coverage in the training corpus, particularly for highly uncommon or out-of-distribution behaviors. Second, the current platform does not include dexterous hand control and is hence unable yet to perform fine-grained dexterous interaction tasks. Extending the framework to humanoids equipped with dexterous hands is an important direction for future work.

\appendix

\section*{\LARGE Appendix}

\section*{Preamble: Shared Setup and Implementation Details}

\paragraph{Training data.}
We adopt the Bones-Seed~\citep{bones_seed} dataset as our motion corpus for training. We deliberately filtered out physically implausible motions, such as in-air stair-climbing or in-air sitting, as well as extreme trajectories such as crawling and jumping, since our focus is on loco-manipulation rather than extreme locomotion. 
All retained sequences are retargeted to the Unitree~G1 skeleton via inverse kinematics.The resulting training corpus contains approximately 140 hours.

\paragraph{Robot platform.}
Experiments use the Unitree G1 humanoid~\citep{unitree_g1} with \textbf{29 actuated joints}:
8~leg joints (hip yaw, hip roll, hip pitch, knee $\times$ 2 sides);
4~ankle joints (ankle pitch, ankle roll $\times$ 2 sides);
3~waist joints (roll, pitch, yaw);
14~arm joints (shoulder pitch/roll/yaw, elbow, wrist roll/pitch/yaw
$\times$ 2 sides).

We define $N=14$ tracking keypoints at anatomically meaningful body locations
(used by the teacher and for reward computation):
pelvis, left/right hip, left/right knee, left/right ankle,
torso, left/right shoulder, left/right elbow, left/right wrist.
For the deployed \emph{partial-keypoint encoder} (Stages~2--3) we use
the five-point sparse subset \textbf{KP5}:
torso, left wrist, right wrist, left ankle, right ankle.

\paragraph{Simulation.}
All training uses \textbf{Isaac Lab}~\citep{nvidia2025isaaclabgpuacceleratedsimulation} with parallel
environments on a mixed flat/rough terrain (50\% flat mesh, 50\% lightly
rough, height noise ${\in}[0.01,0.03]$\,m).
The simulation timestep is $\Delta t_{\text{sim}} = 0.005$\,s (\textbf{200\,Hz});
the policy is queried every four simulation steps (control frequency
\textbf{50\,Hz}, $\Delta t_{\text{ctrl}} = 0.02$\,s).

\section*{A\quad Stage~1: Latent Space and Dynamics-Aware Decoder}

\subsection*{A.1\quad Observations}

Stage~1 trains a privileged \textbf{teacher} tracker with RL, then distills
it online into a deterministic \textbf{encoder–decoder student}
(the joint-command, JC, model).
Table~\ref{tab:s1-obs} itemises both observation spaces ($J=29$).

\begin{table}[h]
\small\centering
\caption{Stage~1 observation spaces.  All teacher terms are stacked over $H{=}5$
history steps; ``dim/step'' is the per-step dimension.}
\label{tab:s1-obs}
\begin{tabular}{@{}llcc@{}}
\toprule
Group & Term & Dim/step & Total \\
\midrule
\multirow{9}{*}{\shortstack[l]{\textbf{Teacher}  \\ (815-dim)}}
  & Reference joint pos.\ + vel.\ & $2J{=}58$ & 290 \\
  & Anchor orientation (6-D, robot-anchor frame)
                                  & 6          & 30  \\
  & Joint positions (rel.\ default) & $J{=}29$ & 145 \\
  & Joint velocities               & $J{=}29$  & 145 \\
  & Base angular velocity          & 3          & 15  \\
  & Last action                    & $J{=}29$  & 145 \\
  & Anchor position (robot-anchor frame)
                                  & 3          & 15  \\
  & Base linear velocity
                                  & 3          & 15  \\
  & Reference base linear velocity (from clip)
                                  & 3          & 15  \\
\midrule
\multirow{7}{*}{\shortstack[l]{\textbf{JC student}  \\ (524-dim)}}
  & $\Delta$ Ref.\ joint pos.\ + vel.\ (goal; current step only)
                                  & $2J{=}58$ & 58  \\
  & Anchor orientation (goal; current step)
                                  & 6          & 6   \\
  & Anchor position (goal; current step)
                                  & 3          & 3   \\
  & Base linear velocity (goal; current step)
                                  & 3          & 3   \\
  & Ref.\ base linear velocity (goal; current step)
                                  & 3          & 3   \\
  & Joint pos.\ / vel.\ / base ang.\ vel.\ / action ($H{=}5$)
                                  & 90         & 450 \\
  & Goal-mask bit                 & 1          & 1   \\
\bottomrule
\end{tabular}
\end{table}

\subsection*{A.2\quad Teacher Reward}

The teacher is trained with PPO using
$r = \sum_i w_i\,r_i$, where $r_i = \exp(-e_i^2/\sigma_i^2)$ for tracking
terms and signed penalties otherwise (Table~\ref{tab:teacher-reward}).

\begin{table}[h]
\small\centering
\caption{Teacher (Stage~1) reward terms.}
\label{tab:teacher-reward}
\begin{tabular}{@{}llcc@{}}
\toprule
Term & Error & $w_i$ & $\sigma_i$ \\
\midrule
\multicolumn{4}{l}{\textit{Kinematic tracking (14 bodies)}} \\
Anchor position         & World-frame pos.          & 2.0  & 0.30 \\
Anchor orientation      & World-frame ori.          & 1.0  & 0.40 \\
Anchor linear velocity  & World-frame vel.          & 3.0  & 1.00 \\
Body position           & Relative pos.\ (14 links) & 1.0  & 0.30 \\
Body orientation        & Relative ori.             & 1.0  & 0.40 \\
Body linear velocity    & Global lin.\ vel.         & 1.5  & 1.00 \\
Body angular velocity   & Global ang.\ vel.         & 1.5  & 3.14 \\
\midrule
\multicolumn{4}{l}{\textit{Teleop extension (upper + lower body)}} \\
Full-body position      & Upper+lower extension     & 1.0  & 0.50 \\
VR 3-point position     & Head + 2 wrists           & 0.5  & 0.50 \\
Feet position           & Foot targets              & 1.0  & 0.50 \\
Full-body orientation   & Extension rotation        & 0.5  & 0.50 \\
Full-body angular vel.\ & Extension ang.\ vel.      & 0.5  & 3.14 \\
Full-body linear vel.\  & Extension lin.\ vel.      & 0.5  & 1.00 \\
\midrule
\multicolumn{4}{l}{\textit{Penalties}} \\
Undesired contacts      & Non-foot/wrist force ${>}1$\,N & $-$0.05 & — \\
Action rate ($\ell_2$)  & Consecutive action diff.       & $-$0.10  & — \\
Joint limit violation   &                                & $-$10.0  & — \\
Joint acceleration ($\ell_2$) &                          & $-2.5\!\times\!10^{-7}$ & — \\
Joint torque ($\ell_2$) &                                & $-1.0\!\times\!10^{-5}$ & — \\
\bottomrule
\end{tabular}
\end{table}

\subsection*{A.3\quad Distillation Objective}

Stage~1 distillation trains the JC encoder $E$ and decoder $D$ online
from teacher rollouts with the loss
\begin{equation}
  \mathcal{L}_1 = \mathcal{L}_{\mathrm{BC}} + \lambda_{\mathrm{sm}}\,\mathcal{L}_{\mathrm{sm}},
  \label{eq:L1}
\end{equation}
where $\mathcal{L}_{\mathrm{BC}}$ is a mean-squared behavioural-cloning loss
between the decoded student action and the teacher action, and
$\mathcal{L}_{\mathrm{sm}}$ is a cosine-smoothness regulariser on the latent
trajectory:
\begin{equation}
  \mathcal{L}_{\mathrm{sm}} = 1 - \cos\!\bigl(\hat{z}_t,\;\hat{z}_{t-1}\bigr).
  \label{eq:Lsm}
\end{equation}
We set $\lambda_{\mathrm{sm}} = 0.1$.

\paragraph{Spherical projection.}
The encoder is \emph{deterministic}—no reparameterisation noise—and its output
is L2-normalised before decoding:
$\hat{z} = E(s_p) / \|E(s_p)\|_2$.
This \emph{spherical projection} constrains all latents to lie on the unit
hypersphere, eliminating the norm-drift failure mode of unconstrained BC
and making $\mathcal{L}_{\mathrm{sm}}$ geometrically well-defined.

\subsection*{A.4\quad Training Details}

\paragraph{Teacher PPO.}
\begin{itemize}
\item Rollout: 24 steps/env; 5 update epochs; 4 mini-batches.
\item Clip $\epsilon{=}0.2$; entropy coeff.\ $0.005$; target KL $0.01$.
\item Learning rate $10^{-3}$ (adaptive); $\gamma{=}0.99$; GAE $\lambda{=}0.95$;
      max grad norm $1.0$; up to 200{,}000 iterations.
\end{itemize}

\paragraph{Distillation.}
Same rollout and optimiser configuration; gradient unrolled over 15 steps;
up to 200{,}000 iterations.

\paragraph{Domain randomisation.}
Applied once at startup: rigid-body friction ${\in}[0.3,1.6]$ (static) and
$[0.3,1.2]$ (dynamic); restitution ${\in}[0.0,0.5]$; joint default position
jitter ${\pm}0.01$\,rad; torso CoM displaced
${\pm}2.5$\,cm ($x$) and ${\pm}5$\,cm ($y,z$).
During rollouts, random push impulses with linear velocity ${\pm}0.5$\,m/s
(horizontal) and ${\pm}0.2$\,m/s (vertical) are applied every $1$--$3$\,s.

\subsection*{A.5\quad Model Size}

\begin{table}[h]
\small\centering
\caption{Stage~1 model components.  The teacher MLP is not deployed.}
\label{tab:s1-size}
\begin{tabular}{@{}ll@{}}
\toprule
Component & Specification \\
\midrule
Teacher actor (MLP) & [1024,\,1024,\,512,\,512,\,256,\,256], ELU \\
Encoder $E$ (Transformer) & 2 layers, $d_{\text{model}}{=}192$, 4 heads, FFN\,768, GELU \\
Latent dim $d_z$ & 16 (unit-norm $\hat{z}$) \\
Decoder $D$ (MLP, input $[\hat{z};\,s_{\text{proprio}}]$)
                           & [1024,\,512,\,256,\,128], ELU \\
Proprio dim (per step)     & $J{+}J{+}3{+}J = 90$;\; $H{=}5 \Rightarrow 450$ total \\
\bottomrule
\end{tabular}
\end{table}

\section*{B\quad Stage~2: Partial Keypoint Encoder}

\subsection*{B.1\quad Observations}

The Stage~2 student receives one \textbf{token per keypoint} from the KP5
body set (torso, left wrist, right wrist, left ankle, right ankle).

\paragraph{Per-keypoint token.}
Each body $b$ contributes a position trajectory in robot-anchor frame sampled
at $T{=}15$ symmetrically log-spaced offsets spanning a $0.5$\,s window.
The raw token is $T{\times}3{=}45$-dimensional.
Invisible (masked) bodies have their token set to \texttt{NaN}; these values
are zeroed before the linear projection and structurally excluded from
attention via a per-body key-padding mask,
so the encoder never attends to hallucinated content.
A per-body binary visibility flag accompanies each token.

\paragraph{Proprioception.}
The student also receives $H{=}5$ history frames of joint positions ($J$),
joint velocities ($J$), base angular velocity (3), and last action ($J$),
totalling $90{\times}5{=}450$\,dim.

\paragraph{Privileged teacher group.}
The frozen Stage-1 JC encoder provides latent supervision targets
$\hat{z}_{\mathrm{JC}}$ from its 524-dim observation
(goal\,73-dim + proprio\,450-dim + mask\,1-dim).

\subsection*{B.2\quad Masking Curriculum}

Training proceeds through three phases (Table~\ref{tab:s2-curriculum}).

\begin{table}[h]
\small\centering
\caption{Stage~2 masking curriculum.
$p_{\text{see}}$ = per-body Bernoulli keep probability.}
\label{tab:s2-curriculum}
\begin{tabular}{@{}clll@{}}
\toprule
Phase & Iterations & Sampling mode & $p_{\text{see}}$ schedule \\
\midrule
1 & $0$–$1{,}000$ & Bernoulli & $1.0$ (held; all KP visible) \\
2 & $1{,}000$–$2{,}000$ & Bernoulli & $1.0\!\to\!0.4$ (linear anneal) \\
3 & $2{,}000$–end & 8-mode demo mix$^{*}$ & $0.4$ (held) \\
\bottomrule
\multicolumn{4}{l}{${}^{*}$Uniform over:
  \textsc{full}, \textsc{vr} (wrists\,+\,torso), \textsc{torso},
  \textsc{left-wrist},}\\
\multicolumn{4}{l}{\phantom{${}^{*}$}
  \textsc{right-wrist}, \textsc{both-wrists}, \textsc{both-ankles},
  \textsc{Bernoulli}.}
\end{tabular}
\end{table}

Phases~1--2 warm up the KP front-end weights (linear projection and
body-ID embeddings, which are freshly initialised) before the encoder
encounters occlusion.
Phase~3 mixes structured semantic modes alongside stochastic Bernoulli masking
so the encoder is in-distribution for all deployment configurations used in
Stage~3 and in the interactive-drag demonstrations.

\subsection*{B.3\quad Distillation Objective}

Stage~2 aligns the KP5 encoder's unit-norm latent to the frozen JC latent
via cosine distance, with a small action-anchor regulariser for coverage:
\begin{equation}
  \mathcal{L}_2
  = \underbrace{1 - \cos\!\bigl(\hat{z}_{\mathrm{KP}},\;\hat{z}_{\mathrm{JC}}\bigr)}_{\text{cosine alignment}}
  + \lambda_a\;\underbrace{\bigl\|D(\hat{z}_{\mathrm{KP}},s_{\mathrm{p}}) - a_{\mathrm{JC}}\bigr\|_2}_{\text{action anchor}},
  \label{eq:L2}
\end{equation}
where $D$ denotes the frozen Stage-1 decoder and $a_{\mathrm{JC}}$ is the JC
teacher action.
We use $\lambda_a = 0.05$.
The decoder $D$ is \textbf{frozen} throughout Stage~2; only the KP front-end
(projection and body-ID embeddings) and the shared transformer backbone train.

\subsection*{B.4\quad Training Details}

Rollout and optimiser configuration are identical to Stage~1 (Section~A.4).
The Stage-1 encoder–decoder checkpoint is loaded as a warm-start; KP front-end
weights are randomly initialised.

\subsection*{B.5\quad Model Size}

\begin{table}[h]
\small\centering
\caption{Stage~2 KP encoder (backbone shared with Stage~1 $E$).}
\label{tab:s2-size}
\begin{tabular}{@{}ll@{}}
\toprule
Component & Specification \\
\midrule
Backbone (shared with $E$) & 2 layers, $d_{\text{model}}{=}192$, 4 heads, FFN\,768, GELU \\
KP bodies $N$              & 5 (torso, L/R wrist, L/R ankle) \\
Time slots $T$             & 15 (sym-sparse log-spaced, 0.5\,s window) \\
KP projection              & Linear $45{\to}d_{\text{model}}$, per body \\
Body-ID embedding          & $N{\times}d_{\text{model}}$ learnable lookup \\
Latent $d_z$               & 16 (unit-norm; frozen after Stage~2) \\
Decoder $D$                & frozen Stage-1 weights \\
\bottomrule
\end{tabular}
\end{table}

\section*{C\quad Stage~3: Task-Specific Latent RL}

\subsection*{C.1\quad Observations}

All task families share the frozen Stage-2 KP5 encoder input, with the
keypoint mask pinned to a single semantic mode per task
(Table~\ref{tab:s3-obs}).

\begin{table}[h]
\small\centering
\caption{Stage~3 policy observations per task family.}
\label{tab:s3-obs}
\begin{tabular}{@{}llll@{}}
\toprule
Task & Visible KP & Mask mode & Additional policy dims \\
\midrule
Omni-directional locomotion & Torso       & \textsc{torso}       & — \\
In-air writing              & Right wrist & \textsc{right-wrist} & — \\
Obstacle reach              & Right wrist & \textsc{right-wrist}
  & Obstacle OBBs: $5{\times}7{=}35$ (appended last) \\
\bottomrule
\end{tabular}
\end{table}

For obstacle reach, each of up to five axis-aligned bounding boxes is
encoded as: centre in robot-anchor frame (3), half-extents (3), and a binary
validity flag (1).
Empty box slots are zeroed and flagged as invalid.
These 35 obstacle dims are appended \emph{after} the standard KP5 encoder
input so the frozen encoder's projection weights are undisturbed;
the corrector $g_\xi$ receives them as separate per-box tokens (Section~C.3).

Each task further uses a task-specific asymmetric \textbf{critic} group
that adds clean proprioception, torso world pose/velocity, the unmasked
reference trajectory for the visible keypoint(s), and—for obstacle
reach—the obstacle OBBs.

\subsection*{C.2\quad Reward Composition}

Table~\ref{tab:s3-reward} lists all reward terms.
Positive terms use $r_i = \exp(-e_i^2/\sigma_i^2)$ on world-frame error;
negative terms are squared-norm penalties.

\begin{table}[h]
\small\centering
\caption{Stage~3 reward composition per task family.}
\label{tab:s3-reward}
\begin{tabular}{@{}lccclc@{}}
\toprule
Term & Loco & Write & Reach & $\sigma$ & Description \\
\midrule
\multicolumn{6}{l}{\textit{Task rewards (world-frame, visible KPs only)}} \\
POI position (fine)   & 3.0  & 12.0 & 12.0 & 0.3$\,{}^*$ & World-frame KP pos error \\
POI position (coarse) & —    & —    & 4.0  & 0.8  & Long-range approach gradient \\
POI linear velocity   & 3.0  & 6.0  & —    & 1.0  & World-frame KP vel error \\
\midrule
\multicolumn{6}{l}{\textit{Stability / constraint}} \\
Global anchor pos.     & 1.0 & —    & —   & 0.3 & Anti-drift trajectory anchor \\
Global anchor ori.     & 1.0 & —    & —   & 0.4 & \\
Upright posture ($\ell_2$) & — & $-$1.0 & — & — & Absolute base-tilt penalty \\
Obstacle keepout       & —   & —    & $-$10.0 & — & OBB penetration penalty \\
\midrule
\multicolumn{6}{l}{\textit{Smoothness / limit penalties (shared across all tasks)}} \\
Undesired contacts & \multicolumn{3}{c}{$-$0.05} & — & Non-foot/wrist contact ${>}$1\,N \\
Action rate ($\ell_2$) & \multicolumn{3}{c}{$-$0.10} & — & \\
Joint limit violation  & \multicolumn{3}{c}{$-$10.0} & — & \\
Joint accel.\ ($\ell_2$) & \multicolumn{3}{c}{$-2.5\!\times\!10^{-7}$} & — & \\
Joint torque ($\ell_2$)  & \multicolumn{3}{c}{$-1\!\times\!10^{-5}$}   & — & \\
\bottomrule
\multicolumn{6}{l}{${}^*$ Fine kernel $\sigma{=}0.4$ for reach (wider goal region); coarse $\sigma{=}0.8$.}
\end{tabular}
\end{table}

\subsection*{C.3\quad Corrector Architecture and RL Training}

\paragraph{Residual corrector $g_\xi$.}
The corrector is a shallow per-body-token transformer that sees the same
tokenised inputs as the frozen Stage-2 encoder (KP tokens, proprioception
frames, the frozen encoder's unit-norm latent $\hat{z}_{\mathrm{enc}}$) and
emits a latent correction $\Delta z$.
The corrected policy mean is
\begin{equation}
  \hat{z}_{\mathrm{policy}}
  = \mathrm{norm}\!\bigl(\hat{z}_{\mathrm{enc}} + \alpha\,g_\xi(\cdot)\bigr),
  \qquad \alpha = 1.0.
  \label{eq:corrector}
\end{equation}
The Stage-2 encoder and Stage-1 decoder are both \textbf{frozen};
only $g_\xi$, the latent log-std parameter, and the value network are updated.

\paragraph{Small-gain initialisation.}
The output projection of $g_\xi$ is Xavier-initialised with gain $0.01$,
forcing $\Delta z \approx 0$ at step~0.
The Stage-3 policy therefore reproduces the distilled Stage-2 behaviour
\emph{exactly} at initialisation, providing a safe start and avoiding the
reward collapse that accompanies random-latent exploration on top of a
pretrained decoder.

Token sequence inside $g_\xi$:
$[\texttt{CLS},\;k_1,\ldots,k_N,\;p_1,\ldots,p_H,\;\hat{z}_{\mathrm{enc}}]$,
where $k_b$ is body $b$'s KP token (masked bodies excluded via key-padding
mask, mirroring Stage~2) and $p_t$ is the $t$-th proprioception frame.
For obstacle reach, up to 5 per-box tokens are appended after
$\hat{z}_{\mathrm{enc}}$; empty boxes are excluded via their validity flag.
The \texttt{CLS} token pool is projected to $\Delta z \in \mathbb{R}^{d_z}$.

\paragraph{PPO configuration.}
\begin{itemize}
\item Rollout: 24 steps/env; 5 update epochs; 4 mini-batches;
      \textbf{8{,}192} parallel environments.
\item Clip $\epsilon{=}0.2$; entropy coeff.\ $0.0$; target KL $0.01$.
\item Learning rate $10^{-4}$ (adaptive); $\gamma{=}0.99$; GAE $\lambda{=}0.95$;
      max grad norm $1.0$.
\item Critic warm-up: actor frozen for the first 100 iterations.
\item Maximum 50{,}000 iterations per task.
\end{itemize}

\paragraph{Success thresholds.}
\textit{Writing}: an episode terminates (with value bootstrapping) when the
right-wrist world-frame RMS error exceeds 0.5\,m after a 60-step (1.2\,s)
grace window; doomed episodes are cut early so rollout time is not wasted.
\textit{Obstacle reach}: success is declared—and the episode bootstrapped—when
the wrist is within 8\,cm of the target for 100 consecutive steps
(${\approx}2\,\text{s}$).

\subsection*{C.4\quad Model Size}
We show the model size of the task-specific RL finetune components in 
Table~\ref{tab:s3-size}.

\begin{table}[htbp]
\small\centering
\caption{Stage~3 module sizes.
The KP5 encoder and decoder are frozen throughout.}
\label{tab:s3-size}
\begin{tabular}{@{}ll@{}}
\toprule
Component & Specification \\
\midrule
Corrector $g_\xi$ (Transformer)
  & 1 layer, $d_{\text{model}}{=}64$, 4 heads, FFN\,128, GELU \\
Output gain (last layer) & 0.01 (small-gain init; $\Delta z{\approx}0$ at step 0) \\
Scale $\alpha$           & 1.0 \\
Initial latent std $\sigma_0$ & 0.1 \\
Value network (MLP)
  & [1024,\,512,\,256,\,128], ELU \\
Frozen KP5 encoder $E$   & Stage-2 checkpoint (not updated) \\
Frozen decoder $D$       & Stage-1 checkpoint (not updated) \\
\bottomrule
\end{tabular}
\end{table}

\section*{D\quad Real-World Deployment Snapshots}

\begin{figure}[h]
  \centering
  \includegraphics[width=\linewidth]{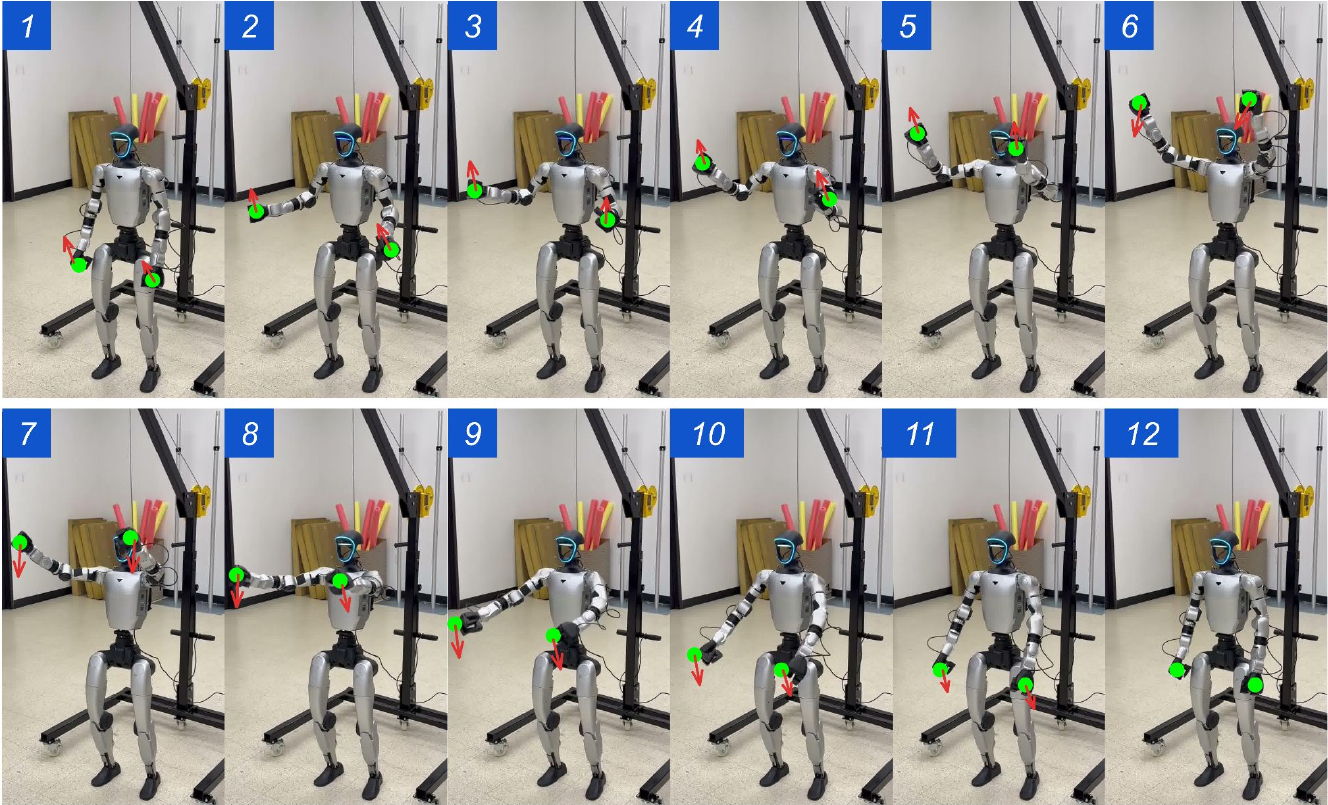}
  \caption{%
    \textbf{Wrist-keypoint following on hardware.}
    Using the wrist keypoints (green dots; red arrows
    show commanded direction), an operator commands the G1 to first raise its arms then lower them. %
  }
  \label{fig:real-drag}
\end{figure}

\begin{figure}[h]
  \centering
  \includegraphics[width=\linewidth]{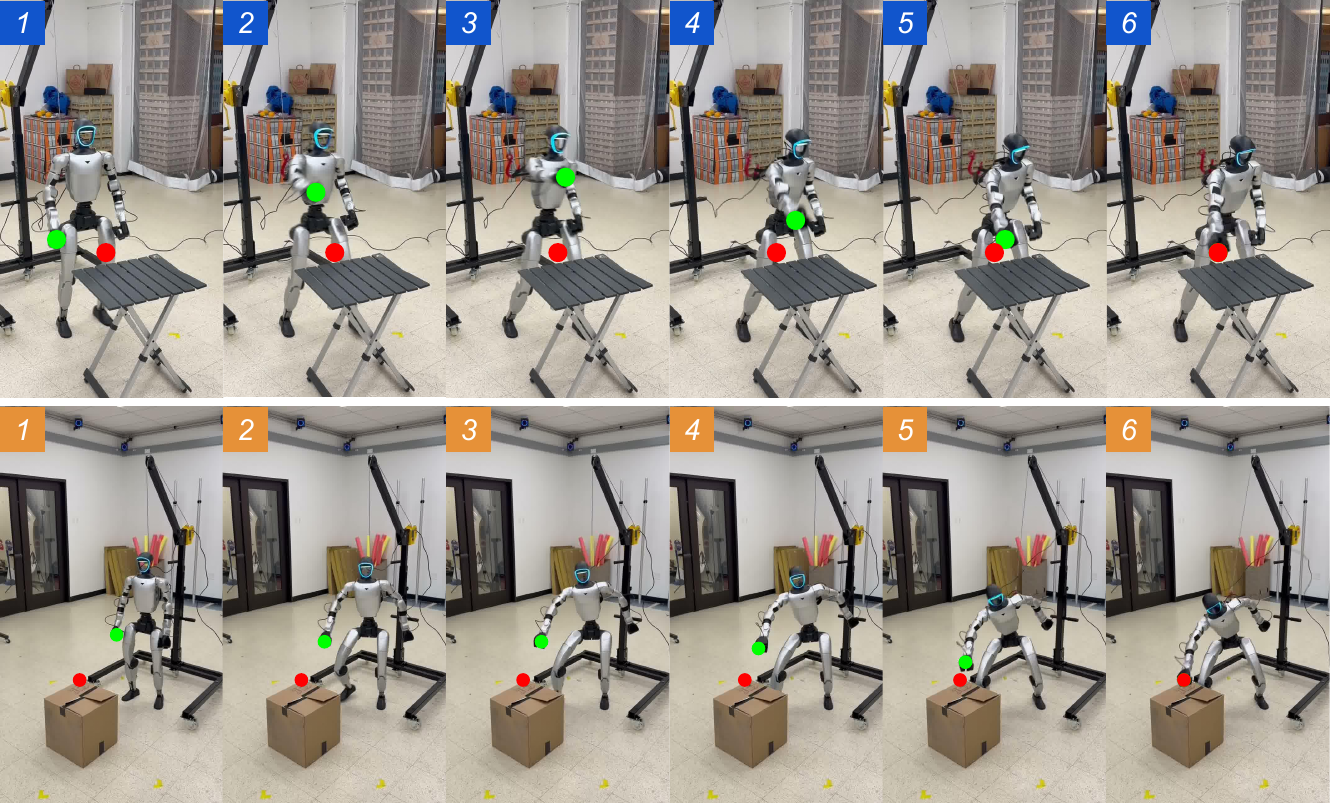}
  \caption{%
    \textbf{Obstacle reach on hardware.}
    The G1 tracks a right-wrist keypoint target (green) while navigating
    around a table (\emph{top}) and a cardboard box requiring a deep-squat
    posture (\emph{bottom}).%
  }
  \label{fig:real-reach}
\end{figure}

\begin{figure}[h]
  \centering
  \includegraphics[width=\linewidth]{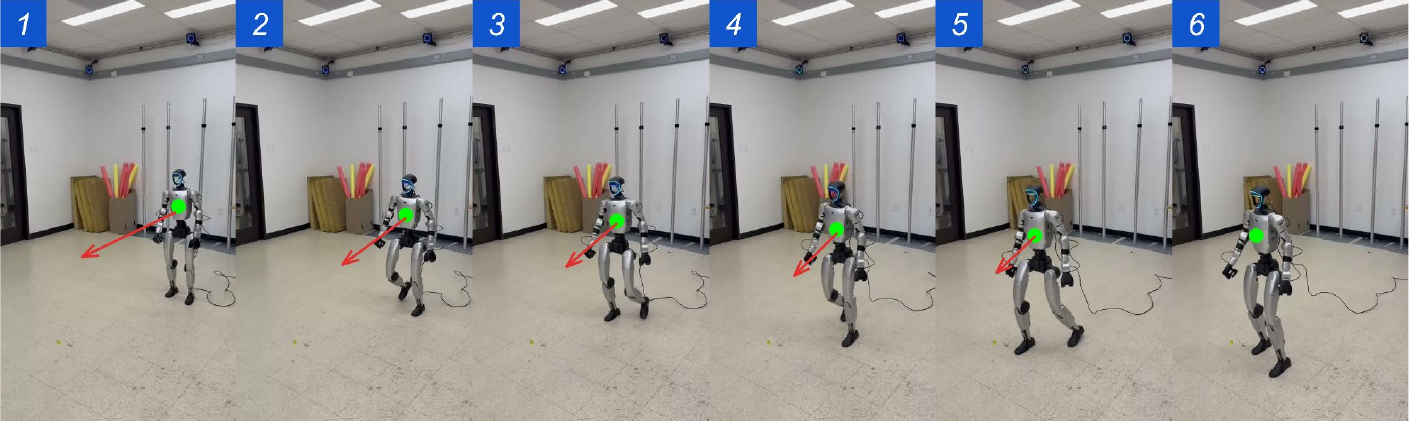}
  \caption{%
    \textbf{Torso-guided locomotion on hardware.}
    Using only the torso keypoint (green dot), an operator commands
    the G1 to walk 3m forward.%
  }
  \label{fig:real-loco}
\end{figure}

We deploy AnyBody zero-shot on a physical Unitree G1.
Figures~\ref{fig:real-drag}--\ref{fig:real-loco} show representative snapshots
from three task families; the green sphere marks the active keypoint target and
the red arrow (or sphere) indicates the operator-commanded trajectory or
goal point.

\clearpage



\bibliography{references}


\end{document}